\documentclass[10pt,twocolumn,letterpaper]{article}

\usepackage{cvpr}
\usepackage{times}
\usepackage{epsfig}
\usepackage{graphicx}
\usepackage{amsmath}
\usepackage{amssymb}
\usepackage{mmstyles}
\usepackage{multirow}
\usepackage{float}
\usepackage{algorithm}  
\usepackage{algorithmicx}
\usepackage{algpseudocode}
\usepackage{bbding}
\usepackage{subcaption}

\usepackage[pagebackref=true,breaklinks=true,letterpaper=true,colorlinks,bookmarks=false]{hyperref}

\cvprfinalcopy 

\def\cvprPaperID{1892} 

\newenvironment{packed_itemize}{
    \vspace{-0.15cm}\begin{itemize}
        \setlength{\itemsep}{1pt}
        \setlength{\parskip}{0pt}
        \setlength{\parsep}{0pt}
    }{\end{itemize}}

\ifcvprfinal\fi

\begin{document}
\title{Deep Flow-Guided Video Inpainting}

\author{Rui Xu$^{1}$ \hspace{9pt} Xiaoxiao Li$^{1}$ \hspace{9pt} Bolei Zhou$^{1}$ \hspace{9pt} Chen Change Loy$^{2}$ \\
	\small{$^{1}$ CUHK-SenseTime Joint Lab, The Chinese University of Hong Kong,
		$^{2}$ Nanyang Technological University} \\ 
	{\tt\small \{xr018, bzhou@ie.cuhk.edu.hk \hspace{5pt} lxx1991@gmail.com \hspace{5pt} ccloy@ntu.edu.sg \}}
 }
\maketitle
\thispagestyle{empty}

\begin{abstract}

Video inpainting, which aims at filling in missing regions of  a video, remains challenging due to the difficulty of preserving the precise spatial and temporal coherence of video contents. In this work we propose a novel flow-guided video inpainting approach. Rather than filling in the RGB pixels of each frame directly, we consider video inpainting as a pixel propagation problem. We first synthesize a spatially and temporally coherent optical flow field across video frames using a newly designed Deep Flow Completion network. Then the synthesized flow field is used to guide the propagation of pixels to fill up the missing regions in the video. Specifically, the Deep Flow Completion network follows a coarse-to-fine refinement to complete the flow fields, while their quality is further improved by hard flow example mining. Following the guide of the completed flow, the missing video regions can be filled up precisely. Our method is evaluated on DAVIS and YouTube-VOS datasets qualitatively and quantitatively, achieving the state-of-the-art performance in terms of inpainting quality and speed. The project page is available at \url{https://nbei.github.io/video-inpainting.html}

\end{abstract}

\section{Introduction}
\label{sec:intro}

\begin{figure*}[t]
	\centering
	\includegraphics[width=0.98\textwidth]{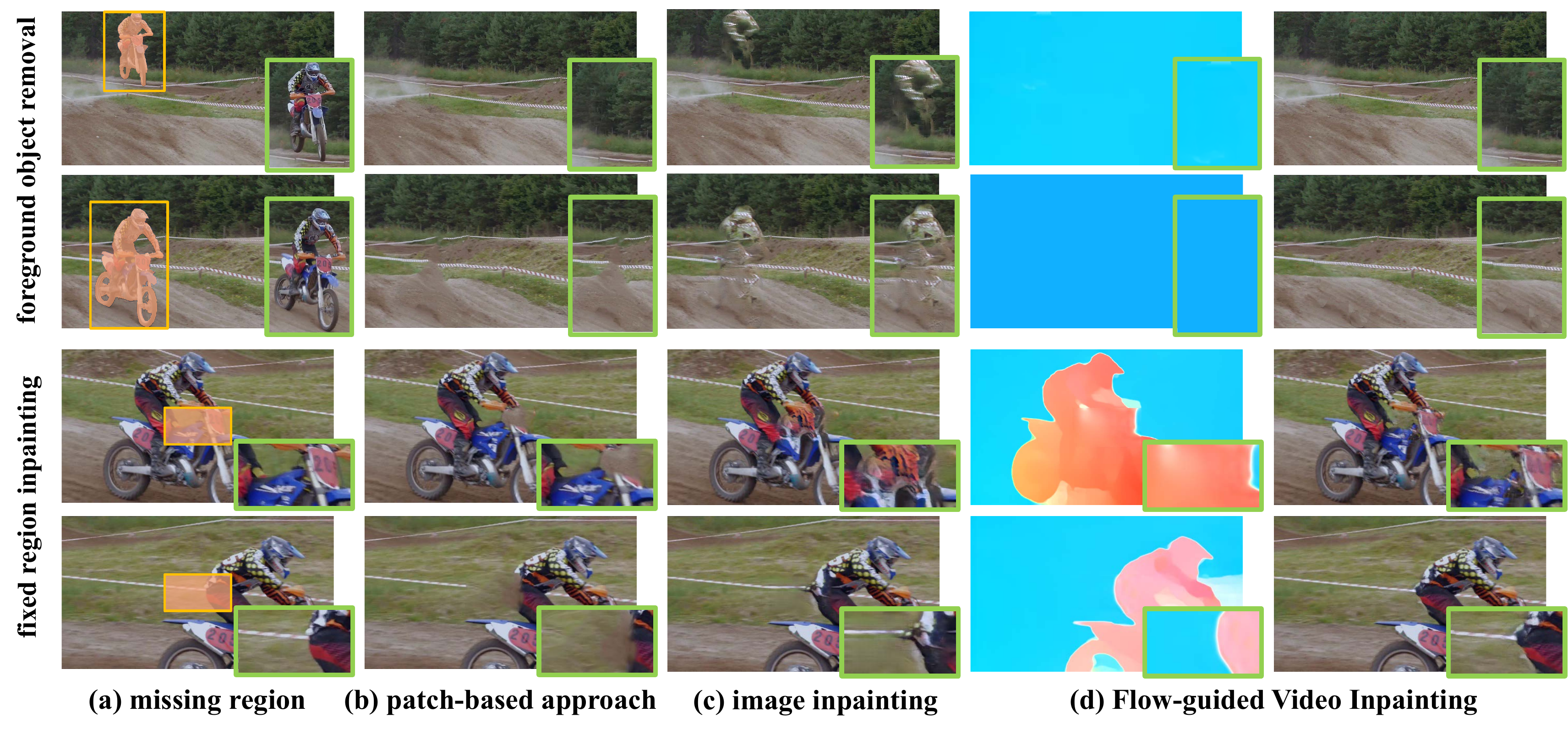}
	\vspace{-10pt}
	\caption{In this example, we show two common inpainting settings, foreground object removal and fixed region inpainting. (a) Missing regions are shown in orange. (b) The result of patch-based optimization approach is affected by complex motions. (c) The image inpainting approach is incapable of maintaining the temporal coherence. (d) Our approach considers the video inpainting as a pixel propagation problem, in which the optical flow field is completed (shown on the left) and then the synthesized flow field is used to guide the propagation of pixels to fill up missing regions (shown on the right). Our inpainting preserves the detail and video coherence.}
	\label{fig:intro}
	\vspace{-14pt}
\end{figure*}

The goal of video inpainting is to fill in missing regions of a given video sequence with contents that are both spatially and temporally coherent~\cite{bertalmio2001navier, Huang-SigAsia-2016, newson2014video, patwardhan2007video}. Video inpainting, also known as video completion, has many real-world applications such as undesired object removal~\cite{ebdelli2015video} and video restoration~\cite{tang2011video}. 

Inpainting real-world high-definition video sequences remains challenging due to the camera motion and the complex movement of objects. 
Most existing video inpainting algorithms~\cite{Huang-SigAsia-2016, newson2013towards, newson2014video, shih2009exemplar, strobel2014flow} follow the traditional image inpainting pipeline, by formulating the problem as a patch-based optimization task, which fills missing regions through sampling spatial or spatial-temporal patches of the known regions then solve minimization problem.
%
%
%
Despite some good results, these approaches suffer from two drawbacks.
First, these methods typically assume smooth and homogeneous motion field in the missing region, therefore they cannot handle videos with complex motions.
%
%
%
%
A failure case is shown in Fig.~\ref{fig:intro}(b).
Second, the computational complexity of optimization-based methods is high thus those methods are infeasible for the real-world applications.
For instance, the method by Huang \etal~\cite{Huang-SigAsia-2016} requires approximately 3 hours to inpaint a 854$\times$480-sized video with 90 frames containing 18\% missing regions.

%
%
Although significant progress has been made in image inpainting~\cite{iizuka2017globally, kohler2014mask, pathak2016context, ren2015shepard, yu2018generative} through the use of Convolutional Neural Network (CNN)~\cite{krizhevsky2012imagenet}, video inpainting using deep learning remains much less explored.
There are several challenges for extending deep learning-based image inpainting approaches to the video domain.
As shown in Fig.~\ref{fig:intro}(c), a direct application of an image inpainting algorithm on each frame individually will lead to temporal artifacts and jitters.
On the other hand, due to the large amount of RGB frames, feeding the entire video sequence at once to a 3D CNN is also difficult to ensure the temporal coherence.
Meanwhile, an extremely large model capacity is needed to directly inpaint the entire video sequence, which is not computationally practical given its large memory consumption.


Rather than filling the RGB pixels, we propose an alternative flow-guided approach for video inpainting.
The motivation behind our approach is that completing a missing flow is much easier than filling in pixels of a missing region directly, while using the flow to propagate pixels temporally preserves the temporal coherence naturally.
As shown in Fig.~\ref{fig:intro}(d), compared with RGB pixels, the optical flow is far less complex and easier to complete since the background and most objects in a scene typically have trackable motion.
This observation inspires us to design our method to alleviate the difficulty of video inpainting by first synthesizing a coherent flow field across frames. 
Most pixels in the missing regions can then be propagated and warped from the visible regions. Finally we can fill up the small amount of regions that are not seen in the entire video using the pixel hallucination~\cite{yu2018generative}. 

%

%
%

In order to fill up the optical flows in videos, we design a  novel \textbf{Deep Flow Completion Network} (DFC-Net) with the following technical novelties:

\noindent \textit{(1) Coarse-to-fine refinement}:
The proposed DFC-Net is designed to recover accurate flow field from missing regions.
This is made possible through stacking three similar subnetworks (DFC-S) to perform coarse-to-fine flow completion.
Specifically, the first subnetwork accepts a batch of consecutive frames as the input and estimates the missing flow of the middle frame on a relatively coarse scale.
The batch of coarsely estimated flow fields is subsequently fed to the second subnetwork followed by the third subnetwork for further spatial resolution and accuracy refinement.

\noindent \textit{(2) Temporal coherence maintenance}:
Our DFC-Net is designed to naturally encourage global temporal consistency even though its subnetworks only predict a single frame each time.
This is achieved through feeding a batch of consecutive frames as inputs, which provide richer temporal information.
In addition, the highly similar inputs between adjacent frames tend to produce continuous results.

 

\noindent \textit{(3) Hard flow example mining}:
We introduce hard flow example mining strategy to improve the inpainting quality on flow boundary and dynamic regions.
%
%
%
%

In summary, the main contribution of this work is a novel flow-guided video inpainting approach. We demonstrate that compelling video completion in complex scenes can be achieved via high-quality flow completion and pixel propagation . A Deep Flow Completion network is designed to cope with arbitrary shape of missing regions, complex motions, and maintain temporal consistency. In comparison to previous methods, our approach is significantly faster in runtime speed, while it does not require any assumptions about the missing regions and the motions of the video contents. We show the effectiveness of our approach on both the DAVIS~\cite{Perazzi2016} and YouTube-VOS~\cite{xu2018youtube} datasets with the state-of-the-art performance.

\section{Related Work}
\label{sec:related_work}

\noindent
\textbf{Non-learning-based Inpainting.}
Prior to the prevalence of deep learning, most image inpainting approaches fall into two categories, \ie, diffusion-based or patch-based methods, which both aim to fill the target holes by borrowing appearance information from known regions.
A diffusion-based method~\cite{ballester2001filling, bertalmio2000image, levin2003learning} propagates appearance information around the target hole for image completion.
This approach is incapable of handling the appearance variations and filling large holes.
A patch-based method~\cite{ bertalmio2003simultaneous, darabi2012image, efros2001image, simakov2008summarizing} completes missing regions by sampling and pasting patches from known regions or other source images.
This kind of approach has been extended to the temporal domain for video inpainting~\cite{newson2013towards, newson2014video, shih2009exemplar}. 
Strobel~\etal~\cite{strobel2014flow} and Huang~\etal~\cite{Huang-SigAsia-2016} further estimate the motion field in the missing regions to address the temporal consistency problem.
In comparison to diffusion-based methods, patch-based methods can better handle non-stationary visual data. However, the dense computation of patch similarity is a very time-consuming operation.
Even by using the PatchMatch~\cite{barnes2009patchmatch, barnes2010generalized} to accelerate the patch matching process, the speed of~\cite{Huang-SigAsia-2016} is still approximately 20 times slower than our approach.
Importantly, unlike our deep learning based approach, all the aforementioned methods cannot capture high-level semantic information. They thus fall short in recovering content in regions that encompasses complex and dynamic motion from multiple objects.   

\begin{figure*}[t]
	\centering
	\includegraphics[width=0.98\textwidth]{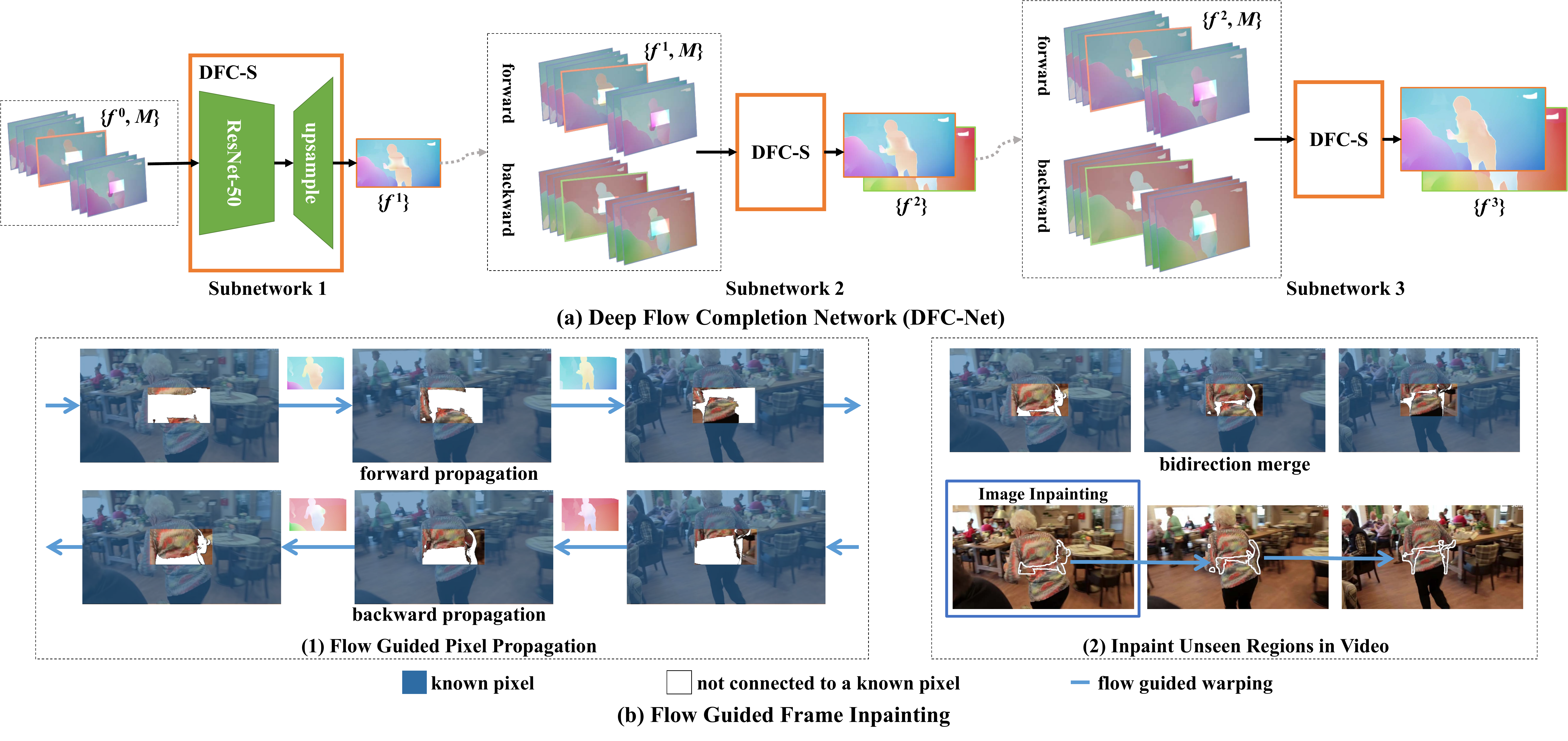}
	\vspace{-10pt}
	\caption{\small{The pipeline of our deep flow-guided video inpainting approach. \textbf{Best viewed with zoom-in.}}}
	\label{fig:pipeline}
	\vspace{-10pt}
\end{figure*}

\noindent
\textbf{Learning-based Inpainting.}
The emergence of deep learning inspires recent works to investigate various deep architectures for image inpainting.
Earlier works~\cite{kohler2014mask, ren2015shepard} attempted to directly train a deep neural network for inpainting. 
%
With the advent of Generative Adversarial Networks (GAN), some studies~\cite{iizuka2017globally, pathak2016context, yu2018generative} formulate inpainting as a conditional image generation problem.
By using GAN, Pathak~\etal~\cite{pathak2016context} train an inpainting network that can handle large-sized holes.
Iizuka~\etal~\cite{iizuka2017globally} improved~\cite{pathak2016context} by introducing both global and local discriminators for deriving the adversarial losses.
More recently, Yu~\etal~\cite{yu2018generative} presented a contextual attention mechanism in a generative inpainting framework, which further improves the inpainting quality.
These methods achieve excellent results in image inpainting. Extending them directly to the video domain is, however, challenging due to the lack of temporal constraints modeling.
%
In this paper we formulate an effective framework that is specially designed to exploit redundant information across video frames. The notion of pixel propagation through deeply estimated flow fields is new in the literature. The proposed techniques, \eg, coarse-to-fine flow completion, maintaining temporal coherence, and hard flow example mining are shown effective in the experiments, outperforming existing optimization-based and deep learning-based methods.


\section{Methodology}
\label{sec:approach}

%

Figure~\ref{fig:pipeline} depicts the pipeline of our flow-guided video inpainting approach.
It contains two steps, the first step is to complete the missing flow while the second step is to propagate pixels with the guidance of completed flow fields.

In the first step, a Deep Flow Completion Network (DFC-Net) is proposed for coarse-to-fine flow completion.
DFC-Net consists of three similar subnetworks named as DFC-S.
The first subnetwork estimates the flow in a relatively coarse scale and feeds them into the second and third subnetwork for further refinement.
In the second step, after the flow is obtained, most of the missing regions can be filled up by pixels in known regions through a flow-guided propagation from different frames.
A conventional image inpainting network~\cite{yu2018generative} is finally employed to complete the remaining regions that are not seen in the entire video.
Thanks to the high-quality estimated flow in the first step, we can easily propagate these image inpainting results to the entire video sequence.

Section~\ref{sec:dfcs} will introduce our basic flow completion subnetwork DFC-S in detail. The stacked flow completion network, DFC-Net, is specified in Sec.~\ref{sec:stacking}. Finally, the RGB pixel propagation procedure will be clarified in Sec.~\ref{sec:propagation}.

\subsection{Deep Flow Completion Subnetwork (DFC-S)}
\label{sec:dfcs}

%

Two types of inputs are provided to the first DFC-S in our network: (i) a concatenation of flow maps from consecutive frames, and (ii) the associated sequence of binary masks, each of which indicating the missing regions of each flow map. 
The output of this DFC-S is the completed flow field of the middle frame. 
In comparison to using a single flow map input, using a sequence of flow maps and the corresponding masks improves the accuracy of flow completion considerably. 

More specifically, suppose $f^0_{i \to (i+1)}$ represents the initial flow between $i$-th and $(i+1)$-th frames and $M_{i \to (i+1)}$ denotes the corresponding indicating mask.
We first extract the flow field using FlowNet 2.0~\cite{ilg2017flownet} and initialize all holes in $f^0_*$ by smoothly interpolating the known values at the boundary inward.
To complete $f^0_{i \to (i+1)}$, the input $\{f^0_{(i-k) \to (i-k+1)},...,f^0_{i \to (i+1)},...,f^0_{(i+k) \to (i+k+1)}\}$ and $\{M_{(i-k)},...,M_{i},...,M_{(i+k)}\}$ are concatenated along the channel dimension and then fed into the first subnetwork, where $k$ denotes the length of consecutive frames.
Generally, $k=5$ is sufficient for the model to acquire related information and feeding more frames do not produce apparent improvement.
With this setting, the number of input channels is $33$ for the first DFC-S (11 flow maps each for the x- and y-direction flows, and 11 binary masks).
For the second and third DFC-S, inputs and outputs are different. Their settings will be discussed in Sec.~\ref{sec:stacking}.
%

As shown in Fig.~\ref{fig:pipeline}(a), considering the tradeoff between model capacity and speed, DFC-S uses the ResNet-50~\cite{he2016deep} as the backbone.
ResNet-50 consists of five blocks named as `conv1', `conv2\_x' to `conv5\_x'.
We modify the input channel of the first convolution in `conv1' to fit the shape of our inputs (\eg, $33$ in the first DFC-S).
To increase the resolution of features, we decrease the convolutional strides and replace convolutions by dilated convolutions from the `conv4\_x' to `conv5\_x' similar to~\cite{chen2014semantic}.
An upsampling module that is composed of three alternating convolution, relu and upsampling layers are then appended to enlarge the prediction.
To project the prediction to the flow field, we remove the last activation function in the upsampling module.

\subsection{Refine Flow by Stacking}
\label{sec:stacking}
Figure~\ref{fig:pipeline}(a) depicts the architecture of DFC-Net, which is constructed by stacking three DFC-S.
Typically, the smaller the hole, the easier the missing flow can be completed, so we first shrink the size of input frames of the first subnetwork to obtain good initial results.
The frames are then gradually enlarged in the second and third subnetwork to capture more details, following a coarse-to-fine refinement paradigm.
Compared with the original size, inputs for three subnetworks are resized as $1/2$, $2/3$ and $1$ respectively. 
 
After obtaining the coarse flow from the first subnetwork, the second subnetwork focuses on further flow refinement. 
To better align the flow field, the forward and backward flows are refined jointly in the second subnetwork.
Suppose $f^1$ is the coarse flow field generated by the first subnetwork.
For each pair of the consecutive frames, $i$-th frame and $\!(i\!+\!1)$-th frame, the second subnetwork takes a sequence of estimated bidirectional flow $\{f^1_{(i-k) \to (i-k+1)},...,f^1_{i \to (i+1)},...,f^1_{(i+k) \to (i+k+1)}\}$ and $\{f^1_{(i-k) \gets (i-k+1)},...,f^1_{i \gets (i+1)},...,f^1_{(i+k) \gets (i+k+1)}\}$  as input and produces refined flows $\{f^2_{i \to (i+1)}, f^2_{i \gets (i+1)}\}$.
Similar to the first subnetwork, binary masks $\{M_{(i-k)},...,M_{i},...,M_{(i+k)}\}$ and $\{M_{(i-k+1)},...,M_{(i+1)},...,M_{(i+k+1)}\}$ are also fed into the second subnetwork to indicate masked regions of the flow field.
The second subnetwork shares the same architecture as the first subnetwork, however, the number of input and output channels is different.

\begin{figure}
	\centering
	\includegraphics[width=0.48\textwidth]{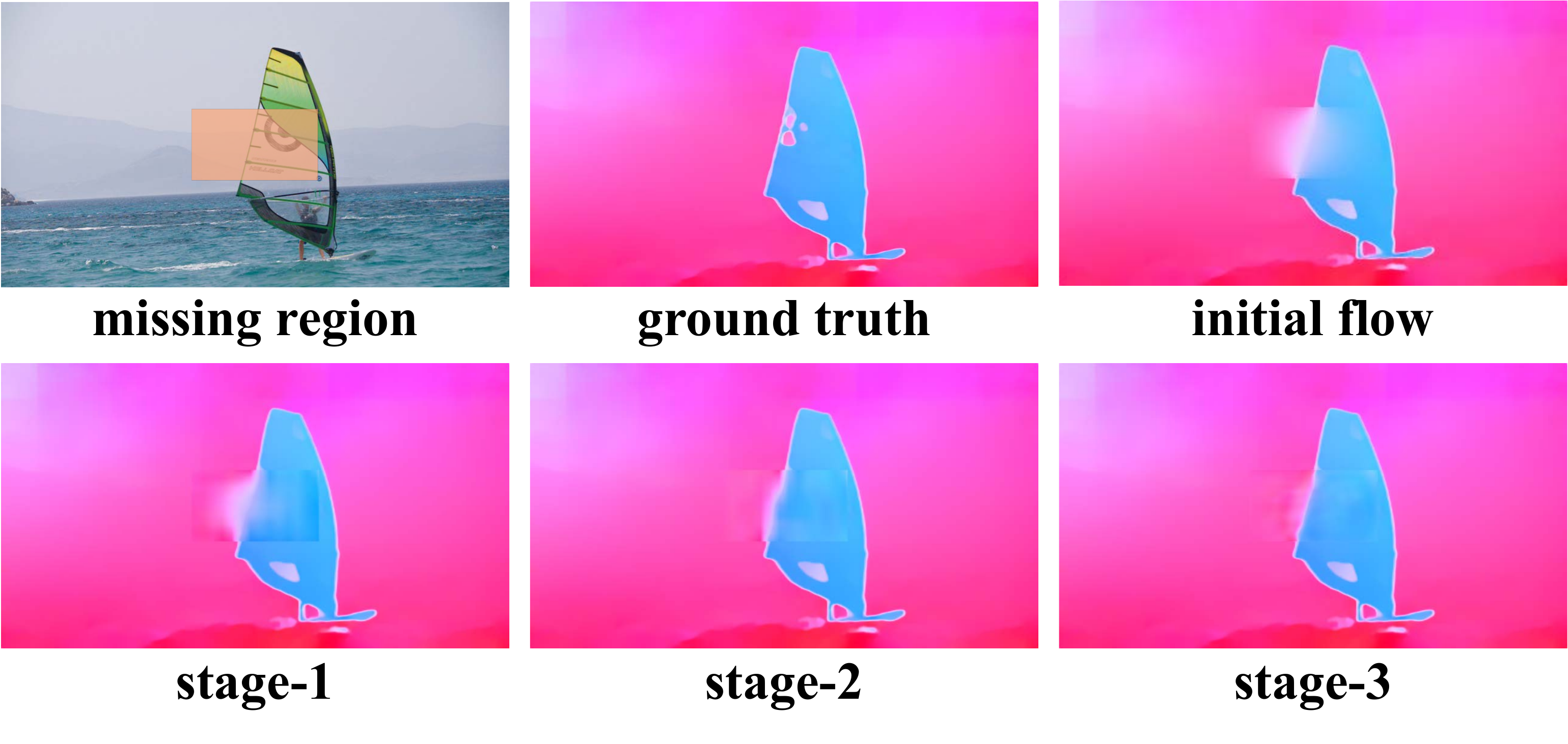}
	\vspace{-20pt}
	\caption{\small{Visualization of different subnetworks’ outputs. The quality of the completed flows is improved over the coarse-to-fine refinement. \textbf{Best viewed with zoom-in.}}}
	\label{fig:stage}
	\vspace{-5pt}
\end{figure}

Finally, predictions from the second subnetwork are enlarged and further fed into the third subnetwork, which strictly follows the same procedure as the second subnetwork to obtain the final results.
A step-by-step visualization is provided in Fig.~\ref{fig:stage}, the quality of the flow field is gradually improved through the coarse-to-fine refinement.

\noindent\textbf{Training.}
During training, for each video sequence, we randomly generate the missing regions. The optimization goal is to minimize the $l_1$ distance between predictions and ground-truth flows.
Three subnetworks are first pre-trained separately and then jointly fine-tuned in end-to-end manner.
Specifically, the loss of the $i$-th subnetwork is defined as:
\vspace{-5pt}
\begin{equation}
{L_i} = \frac{{\Vert M\odot(f^i -\hat{f}) \Vert}_1 }{{\Vert M \Vert}_1},
\end{equation}
where $\hat{f}$ is the ground-truth flow and $\odot$ is element-wise multiplication. For the joint fine-tuning, the overall loss is a linear combination of subnetwork losses.
%
%
%

\noindent\textbf{Hard Flow Example Mining (HFEM).}
Because the majority of the flow area is smooth in video sequences, there exists a huge bias in the number of training samples between the smooth region and the boundary region. In our experiments, we observe that directly using $l_1$ loss generally leads to the imbalanced problem, in which the training process is dominated by smooth areas and the boundary region in the prediction is blurred.
What is worse, the incorrect edge of flow can lead to serious artifacts in the subsequent propagation step.
\begin{figure}
	\centering
	\includegraphics[width=0.48\textwidth]{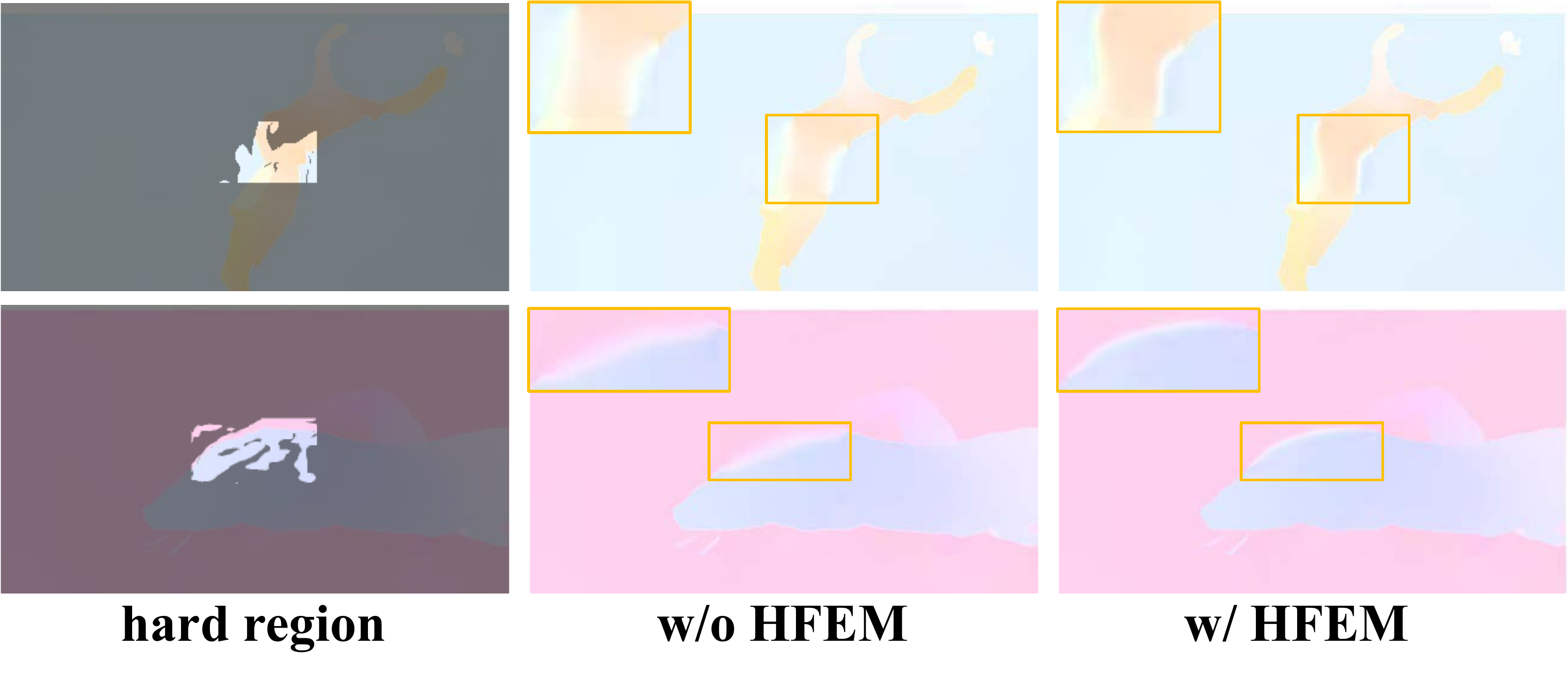}
	\vspace{-20pt}
	\caption{\small{Hard flow example mining.}}
	\label{fig:hfem}
	\vspace{-15pt}
\end{figure}

To overcome this issue, inspired by~\cite{shrivastava2016training},  we leverage the hard flow example mining mechanism to automatically focus more on the difficult areas thus to encourage the model to produce sharp boundaries.
Specifically, we sort all pixels in a descending order of the loss. The top $p$ percent pixels are labeled as hard samples.
Their losses are then enhanced by a weight $\lambda$ to enforce the model to pay more attention to those regions.
The $l_1$ loss with hard flow example mining is defined as:
\vspace{-5pt}
\begin{equation}
{L_i} =  \frac{{\Vert M\odot(f^i -\hat{f}) \Vert}_1 }{{\Vert M \Vert}_1} + \lambda * \frac{{\Vert M^h\odot(f^i -\hat{f}) \Vert}_1 }{{\Vert M^h \Vert}_1},
\end{equation}
where $M^h$ is the binary mask indicating the hard regions. 
As shown in Fig.~\ref{fig:hfem}, the hard examples are mainly distributed around the high frequency regions such as the boundaries.
Thanks to the hard flow example mining, the model learns to focus on producing sharper boundaries.

\subsection{Flow Guided Frame Inpainting}
\label{sec:propagation}

The optical flow generated by DFC-Net establishes a connection between pixels across frames, which could be used as the guidance to inpaint missing regions by propagation.
Figure~\ref{fig:pipeline}(b) illustrates the detailed process of flow-guided frame inpainting .

\noindent
\textbf{Flow Guided Pixel Propagation.}
As the estimated flow may be inaccurate in some locations, we first need to check the validity of the flow.
For a forward flow $f^3_{i \to (i+1)}$ and a location $x_i$, we verify a simple condition based on photometric consistency: ${\Vert (x_{i+1} + f^3_{i \gets (i+1)}(x_{i+1})) - x_i \Vert}_2 < \epsilon,$,
%
where $x_{i+1} = x_i + f^3_{i \to (i+1)}(x_i)$ and $\epsilon$ is a relatively small threshold (\ie, 5).
This condition means that after the forward and backward propagation, the pixel should go back to the original location.
If it is not satisfied, we shall believe that $f^1_{i \to (i+1)}(x_i)$ is unreliable and ignore it in the propagation.
The backward flow can be verified with the same approach.

After the consistency check, as shown in Fig.~\ref{fig:pipeline}(b)(1), all known pixels are propagated bidirectionally to fill the missing regions based on the valid estimated flow.
In particular, if an unknown pixel is connected with both forward and backward known pixels, it will be filled by a linear combination of their pixel values whose weights are inversely proportional to the distance between the unknown pixel and known pixels.

\noindent
\textbf{Inpaint Unseen Regions in Video.}
In some cases, the missing region cannot be filled by the known pixels tracked by optical flow (\eg, white regions in Fig.~\ref{fig:pipeline}(b)(2)), which means that the model fails to connect certain masked regions to any pixels in other frames.
The image inpainting technique~\cite{yu2018generative} is employed to complete such unseen regions.
Figure~\ref{fig:pipeline}(b)(2) illustrates the process of filling unseen regions.
In practice, we pick the a frame with unfilled regions in the video sequence and apply~\cite{yu2018generative} to complete it.
The inpainting result is then propagated to the entire video sequence based on the estimated optical flow.
A single propagation may not fill all missing regions, so image inpainting and propagation steps are applied iteratively until no more unfilled regions can be found.
In average, for a video with 12\% missing regions, there are usually 1\% of unseen pixels and they can be filled after 1.1 iterations.

\section{Experiments}
\label{sec:experiments}
\noindent
\textbf{Inpainting Settings.}
Two common inpainting settings are considered in this paper.
The first setting aims to remove the undesired foreground object, which has been explored in the previous work~\cite{Huang-SigAsia-2016, newson2014video}.
In this setting, a mask is given to outline the region of the foreground object.
In the second setting, we want to fill up an arbitrary region in the video, which might contain either foreground or background. This setting corresponds to some real-world applications such as watermark removal and video restoration.
To simulate this situation, following~\cite{iizuka2017globally, yu2018generative}, a square region in the center of video frames is marked as the missing region to fill up. 
Unless otherwise indicated, for a video frame with size $H \times W$, we fix the size of the square missing region as $H/4 \times W/4$.
The non-foreground mask typically leads to inaccurate flow field estimation, which makes this setting more challenging.

\noindent
\textbf{Datasets.}
To demonstrate the effectiveness and generalization ability of the flow-guided video inpainting approach, we evaluate our method on DAVIS~\cite{Perazzi2016} and YouTube-VOS~\cite{xu2018youtube} datasets.
DAVIS dataset contains 150 high-quality video sequences.
A subset of 90 videos has all frames annotated with the pixel-wise foreground object masks, which is reserved for testing.
For the remaining 60 unlabeled videos, we adopt them for training.
Although DAVIS is not originally proposed for the evaluation of video inpainting algorithms, it is adopted here because of the precise object mask annotations. 
YouTube-VOS~\cite{xu2018youtube} consists of 4,453 videos, which are split into 3,471 for training, 474 for validation and 508 for testing.
Since YouTube-VOS does not provide dense object mask annotations, we only use it to evaluate the performance of the models in second inpainting setting.

\noindent
\textbf{Data Preparation and Evaluation Metric.}
FlowNet 2.0~\cite{ilg2017flownet} is used for flow extraction. The data preparation is different for the two inpainting settings as follows. 

\noindent
(1) Setting 1: foreground object removal.
To prepare the training set, we synthesize and overlay a mask of random shape onto each frame of a video. Random motion is introduced to simulate the actual object mask. Masked and unmasked frames form the training pairs.
For testing, since the ground-truths of removed regions are not available, evaluations are thus conducted through a user study.

\noindent
(2) Setting 2: fixed region inpainting.
Each of the training frame is covered by a fixed square region at the center of the frame. Again, masked and unmasked frames form the training pairs.
For testing, besides the user study, we also report the PSNR and SSIM following~\cite{liu2018image,wang2004image} in this setting. PSNR measures image's distortion, while SSIM measures the similarity in structure between the two images.
%

\subsection{Main Results}
\label{sec:performance}

We quantitatively and qualitatively compare our approach with other existing methods on DAVIS and YouTube-VOS datasets. 
%
For YouTube-VOS, our model is trained on its training set.
The data in DAVIS dataset is insufficient for training a model from scratch. We thus use the pretrained model from YouTube-VOS and fine-tune it using the DAVIS training set.
The performances are reported on their respective test set.

\noindent
\textbf{Quantitative Results.}
We first make comparison with existing methods quantitatively on the second inpainting task that aims to fill up a fixed missing region. The results are summarized in Table~\ref{tab:overall}.

\begin{table}[t]
	\small
	\caption{Quantitative results for the fixed region inpainting.}
	\vspace{-5pt}
	\centering
    \begin{tabular}{@{}l@{\,}|@{}c@{\,}@{}c@{\,}|@{}c@{\,}@{}c@{\,}|@{}c@{\,}}
		& \multicolumn{2}{@{}c@{\,}|}{YouTube-VOS} & \multicolumn{2}{@{}c@{\,}|}{DAVIS} & \multirow{2}{*}{~time\footnotemark(min.)~} \\ \cline{2-5}
		& ~PSNR~            & ~SSIM~           & ~PSNR~         & ~SSIM~        &                              \\ 
		\hline
		Deepfill~\cite{yu2018generative} & 16.68           & 0.15           & 16.47        & 0.14        & \bf{0.3}                          \\
		Newson~\etal~\cite{newson2014video} & 23.92           & 0.37           & 24.72        & 0.43        & $\sim$270                    \\
		Huang~\etal~\cite{Huang-SigAsia-2016}& 26.48           & 0.39           & 27.39        & 0.44        & $\sim$180                    \\
		\hline
		Ours& \bf{27.49}           & \bf{0.41}           & \bf{28.26}        & \bf{0.48}        & \bf{8.5}                         
	\end{tabular}
	\vspace{-10pt}
	\label{tab:overall}
\end{table}
\footnotetext{Following \cite{Huang-SigAsia-2016}, we report the running time on the ``CAMEL'' video in DAVIS dataset. While Newson~\etal~\cite{newson2014video} have not reported the execution time in the paper, we use the similar environment with \cite{Huang-SigAsia-2016} to test their execution time.}

Our approach achieves the best performance on both datasets.
As shown in Table~\ref{tab:overall}, directly applying the image inpainting algorithm~\cite{yu2018generative} on each frame leads to inferior results.
Compared with conventional video inpainting approaches~\cite{Huang-SigAsia-2016, newson2014video}, our approach could better handle videos with complex motions.
Meanwhile, our approach is significantly faster in runtime speed and thus it is more well-suited for real-world applications.

\noindent
\textbf{User study.}
Evaluation metrics in terms of reconstruction errors are not perfect as there are many reasonable solutions for the original video frames. 
Therefore, we perform a user study to quantify the performance of our approach and existing works~\cite{Huang-SigAsia-2016, yu2018generative} for their inpainting quality. 
%
We use the models trained on DAVIS dataset for this experiment. 
Specifically, we randomly choose 15 videos from DAVIS testing set for each participant. The videos are then inpainted by three approaches~(ours, Deepfill~\cite{yu2018generative}, and Huang~\etal~\cite{Huang-SigAsia-2016}) under two different settings.
To better display the details, the video is played at a low frame rate ($5$ FPS).
For each video sample, participants are requested to rank the three inpainting results after the video is played.

\begin{figure}[t]
	\centering
	\includegraphics[width=0.48\textwidth]{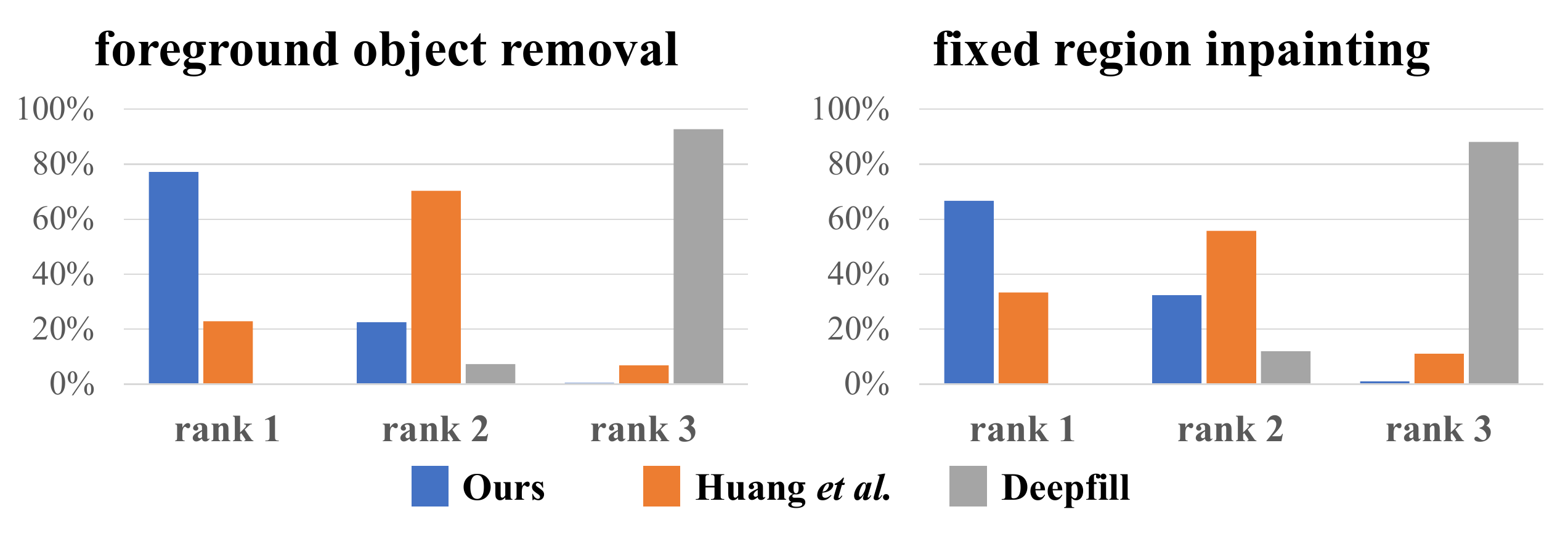}
	\vspace{-16pt}
	\caption{\small{User study. ``Rank x'' means the percentage of inpainting results from each approach being chosen as the x-th best.}}
	\label{fig:user_study}
	\vspace{-8pt}
\end{figure}

\begin{figure*}[t]
	\centering
	\includegraphics[width=0.98\textwidth]{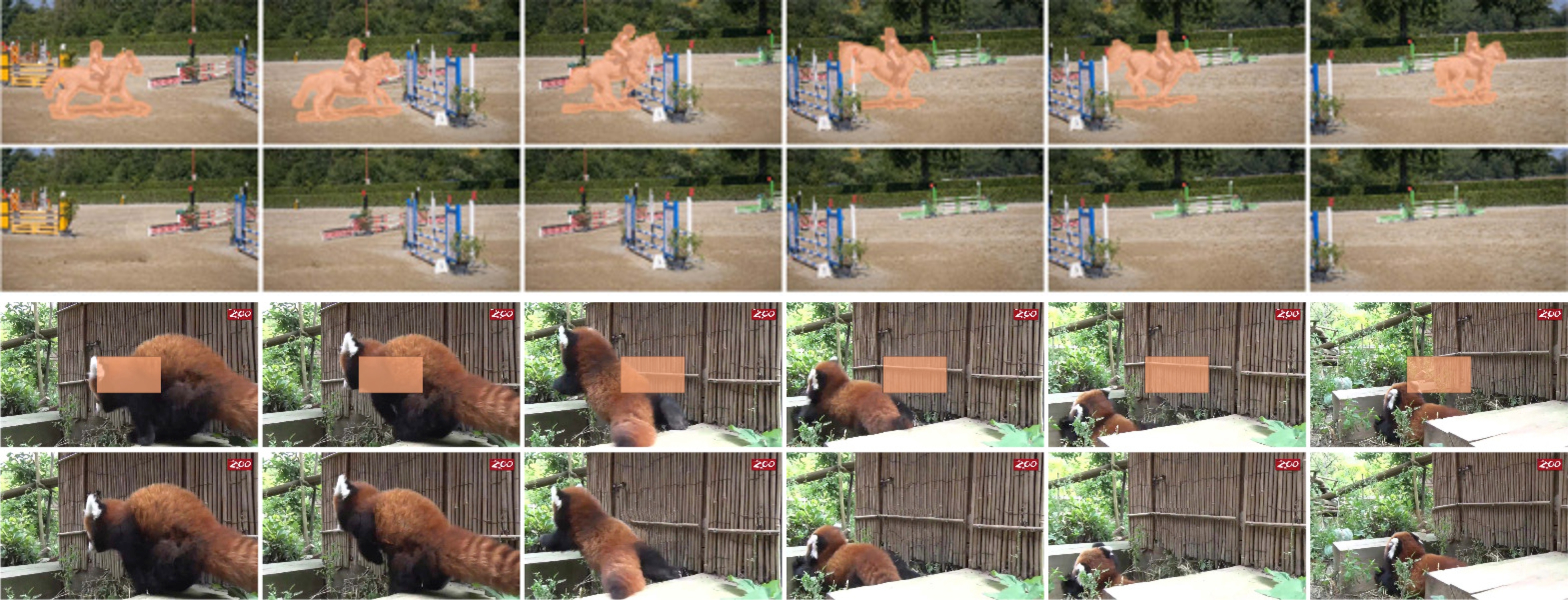}
	\vspace{-5pt}
	\caption{\small{Results of our flow-guided video inpainting approach. For each input sequence (odd row), we show representative frames with mask of missing region overlay. We show the inpainting results in even rows. \textbf{Best viewed with zoom-in.}}
	}
	\label{fig:final}
	\vspace{-5pt}
\end{figure*}

We invited 30 participants for the user study. The result is summarized in Fig.~\ref{fig:user_study}, which is consistent with the quantitative result. Our approach significantly outperforms the other two baselines, while the image inpainting method performs the worst since it is not designed to maintain temporal consistency on its output. Figure~\ref{fig:final} shows some examples of our inpainting results\footnote{We highly recommend watching the video demo in \url{https://youtu.be/zqZjhFxxxus}}.

\begin{figure*}[tb]
	\centering
	\includegraphics[width=0.98\textwidth]{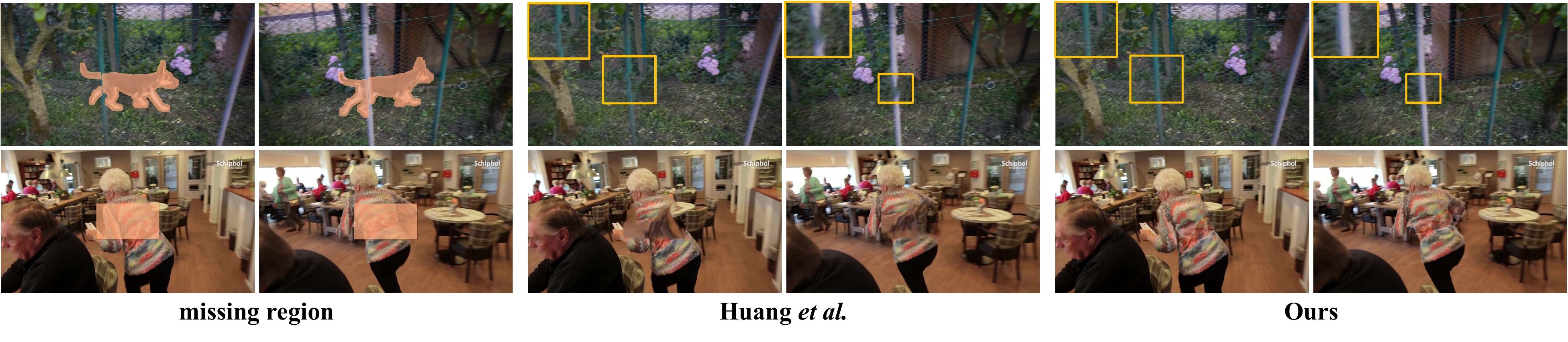}
	\vspace{-13pt}
	\caption{\small{Comparison with Huang~\etal.}}
	\label{fig:visual_compare}
	\vspace{-13pt}
\end{figure*}

\noindent
\textbf{Qualitative Comparison.}
In Fig.~\ref{fig:visual_compare}, we compare our method with Huang~\etal 's method in two different settings.
From the first case, it is evident that our DFC-Net can better complete the flow. 
Thanks to the completed flow, the model can easily fill up the region with correct pixel value.
In the more challenging case shown in the second example,  our method is much more robust on inpainting the complex masked region such as the part of a woman, compared to the notable artifacts in Huang~\etal's result.

\subsection{Ablation Study}
In this section, we conduct a series of ablation studies to analyze the effectiveness of each component in our flow-guided video inpainting approach.
Unless otherwise indicated we employ the training set of YouTube-VOS for training.
For better quantitative comparison, all performances are reported on the validation set of YouTube-VOS under the second inpainting setting, since we have the ground-truth of the removed regions under this setting.

\noindent
\textbf{Comparison with Image Inpainting Approach.}
Our flow-guided video inpainting approach significantly eases the task of video inpainting by using the synthesized flow fields as a guidance, which transforms the video completion problem into a pixel propagation task.
To demonstrate the effectiveness of this paradigm, we compare it with a direct image inpainting network for each individual frame.
%
%
%
%
For a fair comparison, we adopt the Deepfill architecture but with multiple color frames as input, which is named as `Deepfill+Multi-Frame'. 
Then the `Deepfill+Multi-Pass' architecture stacks three `Deepfill+Multi-Frame' like DFC-Net.
Table~\ref{tab:colorbaseline} presents the inpainting results on both DAVIS and YouTube-VOS.
Although the multi-frame input and stacking architecture can bring marginal improvements compared to Deepfill.
The significant gap between `Deepfill+Multi-Frame' and our method demonstrates that using the high-quality completed flow field as guidance can ease the task of video inpainting. 

%
%

\noindent
\textbf{Effectiveness of Hard Flow Example Mining.}
As introduced in Sec.~\ref{sec:stacking}, most of the area of optical flow is smooth and that may result in degenerate models.
Therefore, a hard flow example mining mechanism is proposed to mitigate the influence of the label bias in the problem of flow inpainting.
Similarly, in this experiment, we adopt the first DFC-S to examine the effectiveness of hard flow example mining

	

\begin{table}[t]
	\small
	\caption{Quantitative results for the fixed region inpainting. ``Deepfill+Multi-Frame'' uses Deepfill architecture but with multiple frames as input. ``Deepfill+Multi-Pass'' stacks three ``Deepfill+Multi-Frame'' networks. }
	\centering
	\begin{tabular}{@{}l@{\,}|@{}c@{\,}@{}c@{\,}|@{}c@{\,}@{}c@{\,}}
		& \multicolumn{2}{@{}c@{\,}|}{YouTube-VOS} & \multicolumn{2}{@{}c@{\,}}{DAVIS}  \\ \cline{2-5}
		& ~PSNR~            & ~SSIM~           & ~PSNR~         & ~SSIM~                                     \\ 
		\hline
		Deepfill & 16.68           & 0.15           & 16.47        & 0.14                      \\ 
		Deepfill+Multi-Frame & 16.71           & 0.15           & 16.55        & 0.15                      \\ 
		Deepfill+Multi-Pass & 17.02           & 0.16           & 16.94        & 0.17          \\
		\hline
		Ours & \bf{27.49}           & \bf{0.41}          & \bf{28.26}        & \bf{0.48}        
		
	\end{tabular}
	\vspace{-10pt}
	\label{tab:colorbaseline}
\end{table}


\begin{table}[t]
\small
\caption{Ablation study on hard flow example mining.}
\vspace{-5pt}
\centering
\begin{tabular}{@{}l@{\,}|@{}c@{\,}|@{}c@{\,}|@{}c@{\,}|@{}c@{\,}@{}c@{\,}}
\multirow{2}{*}{$p~(\%)$} & \multicolumn{3}{@{}c@{\,}|}{~Flow completion (EPE)~} & \multicolumn{2}{@{}c@{\,}}{~Video inpainting~} \\ \cline{2-6} 
     &~smooth region~&~hard region~&~overall~&~PSNR~&~SSIM~ \\
\hline\hline
w/o HFEM           & 0.13           & 1.17         & 1.03      & 24.43   & 0.36      \\
70                & 0.13           & 1.13         &    1.01       & 24.63  &   0.36    \\
50                & 0.13           & \bf{1.04}         & 0.99      & \bf{26.15}  &   0.37    \\
30                & 0.13           & \bf{1.04}         &  0.99         & \bf{26.15} &  0.37     \\ 
10                & 0.13           & 1.08         &    1.00       & 25.92   &    0.37  
\end{tabular}
\vspace{-8pt}
\label{tab:ablation_hard}
\end{table}

Table~\ref{tab:ablation_hard} lists the flow completion accuracy under different mining settings, as well as the corresponding inpainting performance.
The parameter $p$ represents the percentage of samples that are labeled as the hard one.
We use the standard end-point-error~(EPE) metric to evaluate our inpainted flow.
For clear demonstration, all flow samples are divided into smooth and non-smooth sets according to their variance.
Overall, the hard flow example mining mechanism improves the performance under all settings.
When $p$ is smaller, which means samples are harder, it will increase the difficulty during training. However, if $p$ is larger, the model would not get much improvement compared with the baseline.
The best choice of $p$ ranges from $30\%$ to $50\%$. In our experiments, we fix $p$ as $50\%$.

\noindent
\textbf{Effectiveness of Stacked Architecture.}
Table~\ref{tab:stacking} depicts the step-by-step refinement results of DFC-Net, including flows and the corresponding inpainting frames.
To further demonstrate the effectiveness of stacked DFC-Net, Table~\ref{tab:stacking} also includes two other baselines that are constructed as follows:

\begin{packed_itemize}
	\vspace{-2pt}
	\small{
		\item \textbf{DFC-Single:}
		DFC-Single is a single stage flow completion network that is similar to DFC-S. To ensure a fair comparison, DFC-Single adopts a deeper backbone, \ie ResNet-101.
		\item \textbf{DFC-Net~(w/o MS):}
		The architecture of DFC-Net~(w/o MS) is the same as DFC-Net. However, in each stage of this baseline model, the input's scale does not change and the data is full resolution from the start to the end.
	}
	\vspace{-5pt}
\end{packed_itemize}

%

\begin{table}[t]

\small
\caption{Ablation study on stacked architecture.}
\vspace{-7pt}
\centering
\begin{tabular}{l|@{}c@{\,}|@{}c@{\,}|@{}c@{\,}}
\multirow{2}{*}{} & ~Flow completion~ & \multicolumn{2}{@{}c@{\,}}{~Video inpainting~}                 \\ \cline{3-4} 
                  & (EPE)                        & ~PSNR~           & ~SSIM~ \\ \hline\hline
Region-Fill       & 1.07                       & 23.85          & 0.35                      \\ 
Stage-1           & 0.99                       & 26.15          & 0.37                      \\ 
Stage-2           & 0.94                       & 27.10          & 0.38                      \\ 
DFC-Single        & 0.97                       & 26.58          & 0.37                      \\ 
DFC-Net (w/o MS)   & 0.95                       & 27.02          & 0.40                      \\ \hline
DFC-Net~(Stage-3) & \textbf{0.93}              & \textbf{27.50} & \textbf{0.41}             \\ 
\end{tabular}
\label{tab:stacking}
\vspace{-8pt}
\end{table}

By inspecting Table~\ref{tab:stacking} closer, we could find that the end-point-error is gradually reduced by the coarse-to-fine refinement.
The result of DFC-Single is somewhat inferior to the second stage, which suggests the effectiveness of using the stacked architecture in this task.
To further indicate the effectiveness of using multi-scale input in each stage, we compare our DFC-Net with DFC-Net~(w/o MS).
The performance gap verifies that the strategy of using multi-scale input in each stage improves the result of our model since using the large scale's input in the early stage typically causes the instability of training.

\noindent
\textbf{Effectiveness of Flow-Guided Pixel Propagation.}
After obtaining the completed flow, all known pixels are first propagated bidirectionally to fill the missing regions based on the valid estimated flow.
This step produces high-quality results and also reduces the size of missing regions that have to be handled in the subsequent step.
\begin{table}[t]
\small
\caption{Ablation study on flow-guided pixel propagation.}
\vspace{-8pt}
\centering
\begin{tabular}{l|c|c}
              & PSNR             & SSIM            \\ \hline \hline
w/o pixel propagation & 19.43          & 0.24          \\ \hline
w/ pixel propagation & \textbf{27.50} & \textbf{0.41}
\end{tabular}
\label{tab:propagation}
\vspace{-10pt}
\end{table}

As shown in Table~\ref{tab:propagation}, compared with a baseline approach that directly use the image inpainting and flow warping to inpaint unseen regions, this intermediate step greatly eases the task and improves the overall performance.

\begin{table}[t]
	\small
	\caption{Ablation study on the quality of the initial flow on DAVIS.}
	\vspace{-10pt}
	\centering
	\begin{tabular}{l|c|c|c}
		
		&EPE     & PSNR    & SSIM   \\ \hline \hline
		Huang~\etal w/o Flownet2 & --  & 27.39 & 0.44 \\ \hline
		Huang~\etal w/ FlowNet2 & 1.02 & 27.73 & 0.45 \\ \hline
		ours           &  \bf{0.93}  & \bf{28.26}  & \bf{0.48}
	\end{tabular}
	\label{tab:initialflow}
	\vspace{-5pt}
\end{table}

\begin{figure}[t]
	\centering
	\includegraphics[width=0.48\textwidth]{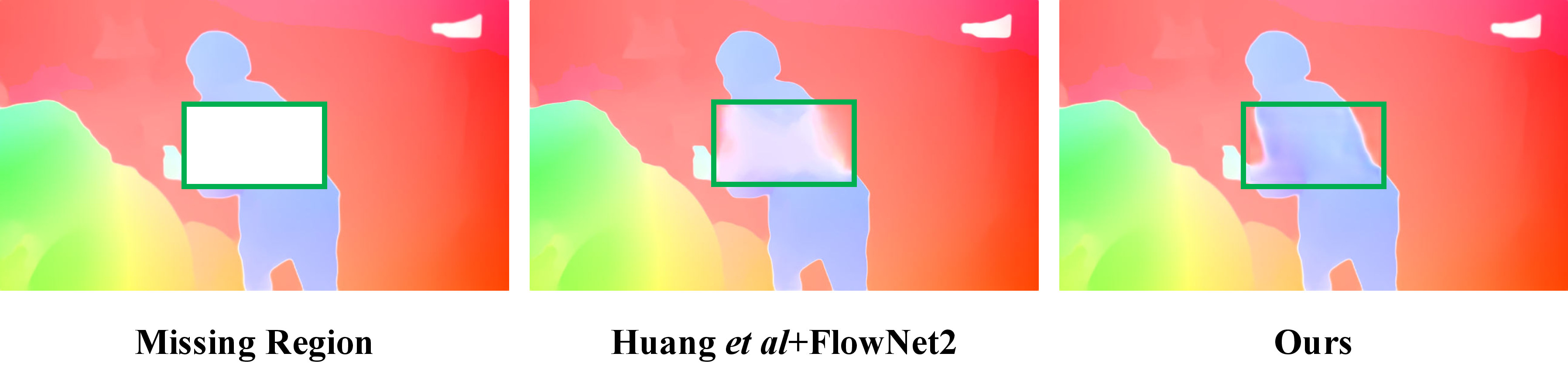}
	\vspace{-15pt}
	\caption{\small{Comparison of completed flow between Huang~\etal and ours.}}
	\label{fig:initialflow}
	\vspace{-13pt}
\end{figure}

\noindent
\textbf{Ablation Study on Initial Flow.}
The flow estimation algorithm is important but not vital since it only affects the flow quality outside the missing regions.
By contrast, the quality of the completed flow inside the missing regions is more crucial.
We substitute the initial flow of \cite{Huang-SigAsia-2016} with flow estimated by FlowNet2 to ensure a fair comparison.
%
%
Table~\ref{tab:initialflow} and Fig.~\ref{fig:initialflow} demonstrate the effectiveness of our method.

\begin{figure}[t]
	\centering
	\includegraphics[width=0.48\textwidth]{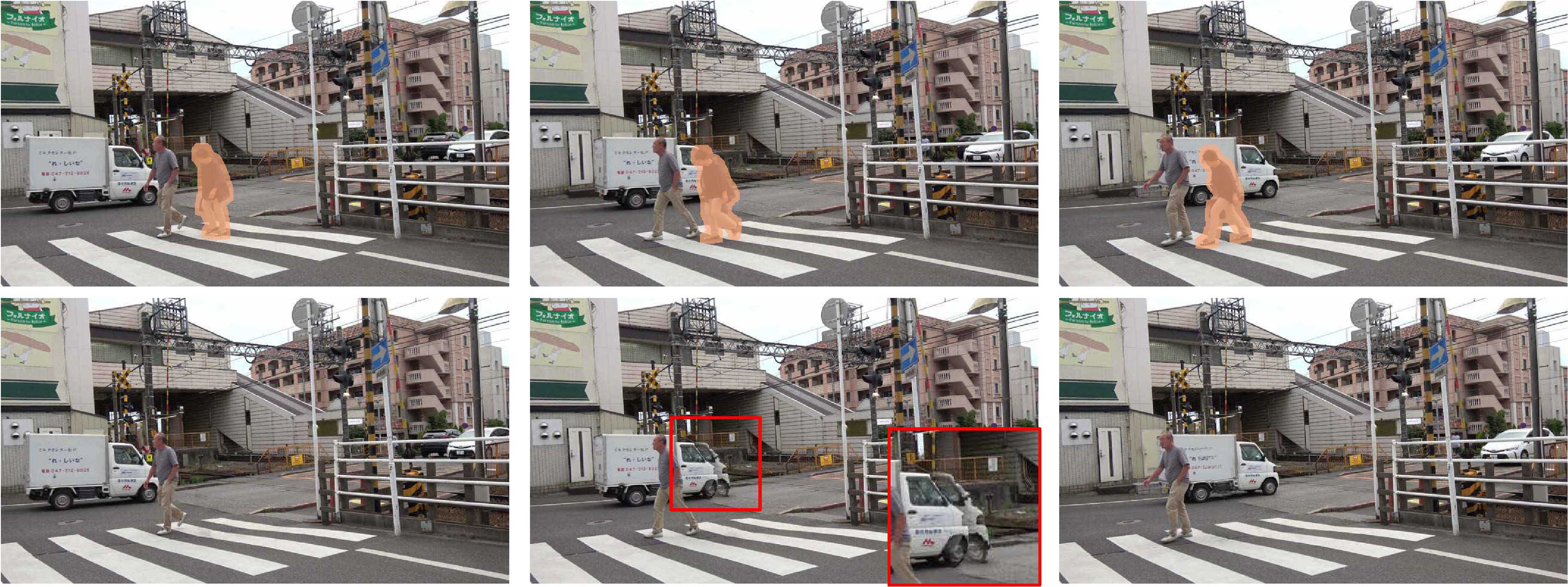}
	\vspace{-10pt}
	\caption{\small{A failure case. The input is shown in the first row, and the output is shown in the second row.}}
	\label{fig:failure}
	\vspace{-12pt}
\end{figure}

\noindent
\textbf{Failure Case.}
A failure case is shown in Fig.~\ref{fig:failure}. 
Our method failed in this case mainly because the completed flow is inaccurate on the edge of the car.
The propagation process cannot amend that.
In the future, we will use the learning based propagation method to mitigate the influence of the inaccuracy of the estimated flow. Other more contemporary flow estimation methods~\cite{hui2018liteflownet,hui2019lightweight,sun2018pwcnet} will be investigated too.

\section{Conclusion}

We propose a novel deep flow-guided video inpainting approach, showing that high-quality flow completion could largely facilitate inpainting videos in complex scenes. Deep Flow Completion network is designed to cope with arbitrary missing regions, complex motions, and yet maintain temporal consistency. In comparison to previous methods, our approach is significantly faster in runtime speed, while it does not require any assumption about the missing regions and the movements of the video contents. We show the effectiveness of our approach on both the DAVIS~\cite{Perazzi2016} and YouTube-VOS~\cite{xu2018youtube} datasets with the state-of-the-art performance.

\vspace{0.1cm}
\noindent
\textbf{Acknowledgements.}
This work is supported by SenseTime Group Limited, the General Research Fund sponsored by the Research Grants Council of the Hong Kong SAR (CUHK 14241716, 14224316. 14209217), and Singapore MOE AcRF Tier 1 (M4012082.020).

\newpage

{\small
\bibliographystyle{ieee}
\bibliography{main_camera}
}

\end{document}